\documentclass[11pt]{article}
\usepackage{acl2016}
\usepackage{pslatex}
\usepackage{latexsym}
\usepackage{amsmath}
\usepackage{amsfonts}
\usepackage{multirow}
\usepackage{url}
\usepackage{booktabs}
\usepackage{tikz}
\usepackage{filecontents}
\usepackage{footnote}
\makesavenoteenv{tabular}
\makesavenoteenv{table}

\let\svtikzpicture\tikzpicture
\def\tikzpicture{\noindent\svtikzpicture}

\usetikzlibrary{%
  arrows,%
  backgrounds,%
  calc,%
  fit,%
  positioning,%
  shadows,%
  shapes,%
  shapes.arrows,%
  shapes.multipart,%
  decorations.pathreplacing,%
  topaths%
}
\tikzstyle{default}=[font=\sffamily,>=stealth']

\tikzset{
  redondo/.style={
    draw=blue,
    line width=1pt,
    rounded corners=3pt,
    text width=#1
  },
  punto/.style={
    fill=red,
    circle,
    inner sep=1.25pt
  },
  cuadra/.style={
    fill=teal,
    minimum size=10pt
  },
  arr/.style={
    line width=1pt,
    draw=green!70!black,
    ->,
    >=latex
  }  
}

\makeatletter
\newcommand{\@BIBLABEL}{\@emptybiblabel}
\newcommand{\@emptybiblabel}[1]{}
\makeatother

\aclfinalcopy

\title{Neural Semantic Role Labeling with Dependency Path Embeddings}

\author{Michael Roth \and Mirella Lapata\\
School of Informatics, University of Edinburgh\\
10 Crichton Street, Edinburgh EH8 9AB\\
\texttt{\{mroth,mlap\}@inf.ed.ac.uk}
}

\date{}

\begin{document}
\maketitle

\begin{abstract}
This paper introduces a novel model for semantic role labeling that makes use of neural sequence modeling techniques. Our approach is motivated by the observation that complex syntactic structures and related phenomena, such as nested subordinations and nominal predicates, are not handled well by existing models. Our model treats such instances as sub-sequences of lexicalized dependency paths and learns suitable embedding representations. We experimentally demonstrate that such embeddings can improve results over previous state-of-the-art semantic role labelers, and showcase qualitative improvements obtained by our method.
\end{abstract}

\section{Introduction}

The goal of \emph{semantic role labeling} (SRL) is to identify and label the arguments of semantic predicates in a sentence according to a set of predefined
relations (e.g.,~``who'' did ``what'' to ``whom''). Semantic roles provide a layer of abstraction beyond syntactic dependency relations, such as \textit{subject} and \textit{object}, in that the provided labels are insensitive to syntactic alternations and can also be applied to nominal predicates. Previous work has shown that semantic roles are  useful for a wide range of natural language processing tasks, with recent applications including statistical machine translation \cite{aziz11,xiong12}, plagiarism detection \cite{osman12,paul15}, and multi-document abstractive summarization \cite{khan15}.

\begin{table}
\begin{center}
\begin{tabular}{@{}c@{ ~}l@{}}
\toprule
System & Analysis \\
\midrule
mate-tools & *He had [trouble$_\text{A0}$] \textbf{raising} [funds$_\text{A1}$]. \\
mateplus  & *He had [trouble$_\text{A0}$] \textbf{raising} [funds$_\text{A1}$]. \\
TensorSRL & *He had trouble \textbf{raising} [funds$_\text{A1}$]. \\
easySRL & *He had trouble \textbf{raising} [funds$_\text{A1}$]. \\
\midrule
This work & [He$_\text{A0}$] had trouble \textbf{raising} [funds$_\text{A1}$]. \\
\bottomrule
\end{tabular}
\end{center}
\label{tbl:johnmary}
\caption{Outputs of SRL systems for the sentence \textsl{He had trouble raising funds}.\ Arguments of \textbf{raise} are shown with predicted roles as defined in PropBank\ (A0:\ getter of money;\ A1:\ money).\ Asterisks mark flawed analyses that miss the argument \textsl{He}.
}
\end{table}

The task of semantic role labeling (SRL) was pioneered by \newcite{gildea02}. In their work, features based on syntactic constituent trees were identified as most valuable for labeling predicate-argument relationships. Later work confirmed the importance of syntactic parse features \cite{pradhan05,punyakanok08} and found that dependency parse trees provide a better form of representation to assign role labels to arguments \cite{johansson08}.

Most semantic role labeling approaches to date rely heavily on lexical and syntactic indicator features. 
Through the availability of large annotated resources, such as PropBank \cite{palmer05}, statistical models based on such features achieve high accuracy. However, results often fall short when the input to be labeled involves instances of linguistic phenomena that are relevant for the labeling decision but appear infrequently at training time. Examples include control and raising verbs, nested conjunctions or other recursive structures, as well as rare nominal predicates. The difficulty lies in that simple lexical and syntactic indicator features are not able to model interactions triggered by such phenomena. For instance, consider the sentence \textsl{He had trouble raising funds} and the analyses provided by four publicly available tools in Table~\ref{tbl:johnmary} (mate-tools, \newcite{bjoerkelund10}; mateplus, \newcite{rothwoodsend14}; TensorSRL, \newcite{lei15}; and easySRL, \newcite{lewis15}). Despite all systems claiming state-of-the-art or competitive performance, none of them is able to correctly identify \textsl{He} as the agent argument of the predicate \textsl{raise}. Given the complex dependency path relation between the predicate and its argument, none of the systems actually identifies \textsl{He} as an argument at all.

In this paper, we develop a new neural network model that can be applied to the task of semantic role labeling. The goal of this model is to better handle control predicates and other phenomena that can be observed from the dependency structure of a sentence. In particular, we aim to model the semantic relationships between a predicate and its arguments by analyzing the dependency path between the predicate word and each argument head word. We consider lexicalized paths, which we decompose into sequences of individual items, namely the words and dependency relations on a path. We then apply long-short term memory networks \cite{hochreiter97} to find a recurrent composition function that can reconstruct an appropriate representation of the full path from its individual parts (Section~\ref{sec:depemb}). To ensure that representations are indicative of semantic relationships, we use semantic roles as target labels in a supervised setting (Section~\ref{sec:model}). 

By modeling dependency paths as sequences of words and dependencies, we implicitly address the data sparsity problem. This is the case because we use single words and individual dependency relations as the basic units of our model. In contrast, previous SRL work only considered full syntactic paths. 
Experiments on the CoNLL-2009 benchmark dataset show that our model is able to outperform the state-of-the-art in English 
(Section~\ref{sec:exp}), and that it improves SRL performance in other languages, including Chinese, German and Spanish~(Section~\ref{sec:multiling}).

\section{Dependency Path Embeddings}
\label{sec:depemb}

In the context of neural networks, the term \textit{embedding} refers to the output of a function $f$ within the network, which transforms an arbitrary input into a real-valued vector output. Word embeddings, for instance, are typically computed by forwarding a one-hot word vector representation from the input layer of a neural network to its first hidden layer, usually by means of matrix multiplication and an optional non-linear function whose parameters are learned during neural network training.

\begin{figure}[t]
%\begin{center}
%\includegraphics[scale=0.7]{network.jpg}
\fbox{\vbox{
\begin{tikzpicture}[
  %->,
  >=stealth',
  %shorten >=1pt,
  auto,
  node distance=2.5cm,
  bend angle=45,
  %thick
  ]
  \node at (-0.4,0) {};
  \node[] (root) at (1.4,1.5) {\small{ROOT}};
  \node[draw,dotted] (he) at (0,0) {he};
  \node () at (0.0,-0.5) {\small{A0}};
  \node[draw,dotted] (had) at (1.4,0.5) {had};
  \node[draw,dotted] (trouble) at (3,0) {trouble};
  \node[draw,dotted] (raising) at (4.9,-0.5) {\textbf{raising}};
  \node () at (4.9,-1) {\small{raise.01}};
  \node (funds) at (6.5,-1) {funds};
  \node () at (6.5,-1.4) {\small{A1}};
    
  \draw[->] (had.north)--(root.south);
  \draw[->] ([yshift=0.2cm]he.east)--node {\small{SBJ}} ([yshift=0.1cm]had.west);  
  \draw[<-] ([yshift=0.1cm]had.east)--node {\small{OBJ}}  ([yshift=0.2cm]trouble.west);
  \draw[<-] ([yshift=0.1cm]trouble.east)--node {\small{NMOD}} ([yshift=0.2cm]raising.west); 
  \draw[<-] ([yshift=0.1cm]raising.east)--node {\small{OBJ}} ([yshift=0.2cm]funds.west); 
  
  \draw[dotted,->] ([yshift=-0.0cm]he.east)--([yshift=-0.1cm]had.west);  
  \draw[dotted,->] ([yshift=-0.0cm]trouble.west)--([yshift=-0.1cm]had.east);  
  \draw[dotted,->] ([yshift=-0.0cm]raising.west)--([yshift=-0.1cm]trouble.east); 
  
\end{tikzpicture}
}}
%\end{center}
\caption{
Dependency path (dotted) between the predicate \textsl{raising} and the argument \textsl{he}.}
\label{fig:deptree}
\vspace{0.5em}
\end{figure}
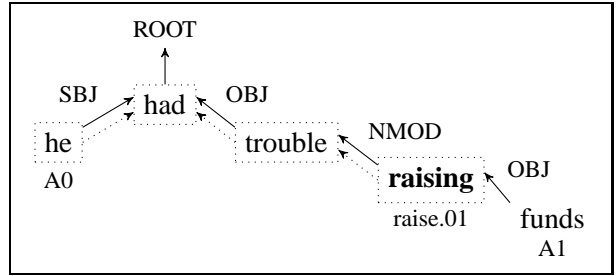

Here, we seek to compute real-valued vector representations for
\textit{dependency paths} between a pair of words $\langle w_i,w_j
\rangle$. We define a dependency path to be the \textit{sequence} of
nodes (representing words) and edges (representing relations between
words) to be traversed on a dependency parse tree to get from node
$w_i$ to node $w_j$. In the example in Figure~\ref{fig:deptree}, the
dependency path from \textsl{raising} to \textsl{he} is
$\textsl{raising}\!  \xrightarrow{_{\text{NMOD}}}\!
\textsl{trouble}\!  \xrightarrow{_{\text{OBJ}}}\!  \textsl{had}\!
\xleftarrow{_{\text{SBJ}}}\!  \textsl{he}$.

Analogously to how word embeddings are computed, the simplest way to embed paths would be to represent each sequence as a one-hot vector. However, this is suboptimal for two reasons: Firstly, we expect only a subset of dependency paths to be attested frequently in our data and therefore many paths will be too sparse to learn reliable embeddings for them. Secondly, we hypothesize that dependency paths which share the same words, word categories or dependency relations should impact SRL decisions in similar ways. Thus, the words and relations on the path should drive representation learning, rather than the full path on its own.
The following sections describe how we address  representation learning by means of modeling dependency paths as sequences of items in a recurrent neural network.

\subsection{Recurrent Neural Networks}

The recurrent model we use in this work is a variant of the long-short term memory (LSTM) network. It takes a sequence of items \mbox{$X={x_1,...,x_n}$} as input, recurrently processes each item \mbox{$x_t\in X$} at a time, and finally returns one embedding state~\textbf{e}$_n$ for the complete input sequence. For each time step $t$, the LSTM model updates an internal memory state~\textbf{m}$_t$ that depends on the current input as well as the previous memory state~\textbf{m}$_{t-1}$. In order to capture long-term dependencies, a so-called gating mechanism controls the extent to which each component of a memory cell state will be modified. In this work, we employ input gates~\textbf{i}, output gates~\textbf{o} and (optional) forget gates~\textbf{f}. We formalize the state of the network at each time step~$t$ as follows:

%%%\vspace{-1.0em}
\begin{gather}
\textbf{i}_t = \sigma([\textbf{W}^\text{mi}\textbf{m}_{t-1}] + \textbf{W}^\text{xi}x_t + \textbf{b}^\text{i}) \\
\textbf{f}_t = \sigma([\textbf{W}^\text{mf}\textbf{m}_{t-1}] + \textbf{W}^\text{xf}x_t + \textbf{b}^\text{f}) \\
\textbf{m}_t = \textbf{i}_t \odot (\textbf{W}^\text{xm}x_t) + \textbf{f}_t \odot \textbf{m}_{t-1} + \textbf{b}^\text{m}\\
\textbf{o}_t = \sigma([\textbf{W}^\text{mo}\textbf{m}_t] + \textbf{W}^\text{xo}x_t + \textbf{b}^\text{o}) \\
\textbf{e}_t = \textbf{o}_t \odot \sigma (\textbf{m}_t)
\end{gather}
\vspace{0.5em}

In each equation, \textbf{W} describes a matrix of weights to project information between two layers, \textbf{b} is a layer-specific vector of bias terms, and $\sigma$ is the logistic function. Superscripts indicate the corresponding layers or gates. Some models described in Section~\ref{sec:model} do not make use of forget gates or memory-to-gate connections. In case no forget gate is used, we set $\textbf{f}_t=\textbf{1}$. If no memory-to-gate connections are used, the terms in square brackets in (1), (2), and (4) are replaced by zeros.

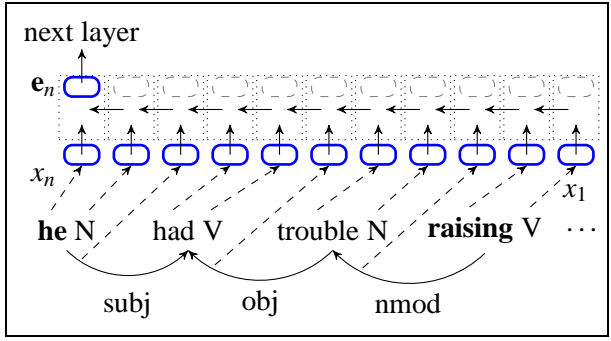
\begin{figure}[t]
\fbox{\vbox{
\begin{tikzpicture}[
  %->,
  >=stealth',
  %shorten >=1pt,
  auto,
  node distance=2.5cm,
  bend angle=45,
  %thick
  ]

\node (he) at (0,0) {\textbf{he} N};
\node (had) at (1.6,0) {had V };
\node (trouble) at (3.5,0) {trouble N};
\node (raise) at (5.5,0) {\textbf{raising} V };
\node (fund) at (6.8,0) {\ldots};
%\node (fund) at (6.3,0) {funds N};

\draw[->] (he.south) to[bend right] node[yshift=-0.7cm] (tmp3) {subj} (had.south);
\draw[->] (trouble.south) to[bend left] node[] (tmp2) {obj} (had.south);
\draw[->] (raise.south) to[bend left] node[] (tmp1) {nmod} (trouble.south);
%\draw[->] (fund.south) to[bend left] node[]{obj} (raise.south);

\node[redondo=0.2cm] (A) at (0.2,1) {};
\node[redondo=0.2cm] (B) at (0.85,1) {};
\node[redondo=0.2cm] (C) at (1.5,1) {};
\node[redondo=0.2cm] (D) at (2.15,1) {};
\node[redondo=0.2cm] (E) at (2.8,1) {};
\node[redondo=0.2cm] (F) at (3.45,1) {};
\node[redondo=0.2cm] (G) at (4.1,1) {};
\node[redondo=0.2cm] (H) at (4.75,1) {};
\node[redondo=0.2cm] (I) at (5.4,1) {};
\node[redondo=0.2cm] (J) at (6.05,1) {};
\node[redondo=0.2cm] (K) at (6.7,1) {};
\node[below of=K, node distance=0.5cm] (K.text) {$x_1$};
\node[] (A.text) at (-0.3,0.7) {$x_n$};

\draw[->,dashed] ([xshift=+0.6cm]raise.north)--(K.south);
\draw[->,dashed] ([xshift=-0.2cm]raise.north)--(J.south);
\draw[->,dashed] ([yshift=+0.2cm,xshift=-0.6cm]tmp1.north)--(I.south);

\draw[->,dashed] ([xshift=+0.6cm]trouble.north)--(H.south);
\draw[->,dashed] ([xshift=-0.2cm]trouble.north)--(G.south);
\draw[->,dashed] ([yshift=+0.2cm,xshift=-0.6cm]tmp2.north)--(F.south);

\draw[->,dashed] ([xshift=+0.3cm]had.north)--(E.south);
\draw[->,dashed] ([xshift=-0.2cm]had.north)--(D.south);
\draw[->,dashed] ([yshift=+0.3cm,xshift=-0.6cm]tmp3.north)--(C.south);

\draw[->,dashed] ([xshift=+0.3cm]he.north)--(B.south);
\draw[->,dashed] ([xshift=-0.2cm]he.north)--(A.south);

\foreach \x in {6.7,6.05,...,0.2}
  \draw[dotted] (\x-0.3, 1.2) rectangle (\x+0.3, 2.05);
\foreach \x in {6.7,6.05,...,0.2}
 \draw[->] (\x,1.0)--(\x,1.4);
\foreach \x in {6.7,6.05,...,0.85}
 \draw[->] (\x-0.1,1.6)--(\x-0.55,1.6);

%\draw[->] (0.2,1.9)--(0.2,2.2);
\foreach \x in {6.7,6.05,...,0.2}
  \node[redondo=0.2cm,dashed,gray,thin] (O) at (\x,1.9) {};
\node[redondo=0.2cm] (O) at (0.2,1.9) {};
\node[] at (-0.3,1.9) {\textbf{e}$_n$};
\draw[->] (0.2,1.95)--(0.2,2.4);
\node[] at (0.2,2.6) {next layer};

\end{tikzpicture}
}}
\caption{Example input and embedding computation for the path from  
\textsl{raising} to \textsl{he}, given the sentence \textsl{he had trouble raising funds}. LSTM time steps are displayed from right to left.}
\label{fig:pathexample}
%\end{center}
\end{figure}

\subsection{Embedding Dependency Paths}

We define the \textit{embedding of a dependency path} to be the final memory output state of a recurrent LSTM layer that takes a path as input, with each input step representing a binary indicator for a part-of-speech tag, a word form, or a dependency relation.
In the context of semantic role labeling, we define each path as a sequence from a predicate to its potential argument.\footnote{We experimented with different sequential orders and found this to lead to the best validation set results.} Specifically, we define the first item $x_1$ to correspond to the part-of-speech tag of the predicate word $w_i$, followed by its actual word form, and the relation to the next word $w_{i+1}$. The embedding of a dependency path corresponds to the state \textbf{e}$_n$ returned by the LSTM layer after the input of the last item, $x_n$, which corresponds to the word form of the argument head word $w_j$. An example is shown in Figure~\ref{fig:pathexample}.

The main idea of this model and representation is that word forms, word categories and dependency relations can all influence role labeling decisions. The word category and word form of the predicate first determine which roles are plausible and what kinds of path configurations are to be expected. The relations and words seen on the path can then manipulate these expectations. In Figure~\ref{fig:pathexample}, for instance, the verb \textsl{raising} complements the phrase \textsl{had trouble}, which makes it likely that the subject \textsl{he} is also the logical subject of \textsl{raising}.

By using word forms, categories and dependency relations as input items, we ensure that specific words (e.g., those which are part of complex predicates) as well as various relation types (e.g., subject and object) can appropriately influence the representation of a path. While learning corresponding interactions, the network is also able to determine which phrases and dependency relations might not influence a role assignment decision (e.g., coordinations).

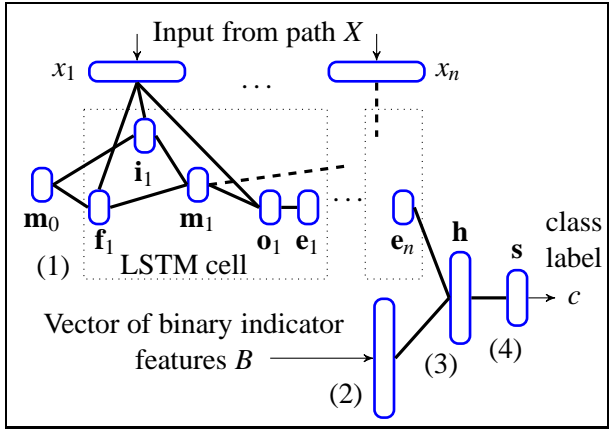
\begin{figure}[t]
\fbox{\vbox{
\begin{tikzpicture}[
  %->,
  >=stealth',
  %shorten >=1pt,
  auto,
  node distance=2.5cm,
  bend angle=45,
  %thick
  ]

\node[] at (-0.6,-0.1) {(1)};
\draw[dotted] (-0.2,2)--(3,2)--(3,-0.25)--(-0.2,-0.25)--(-0.2,2);
\draw[dotted] (3.5,2)--(4.25,2)--(4.25,-0.25)--(3.5,-0.25)--(3.5,2);
\node[redondo=1cm,label={left:$x_1$}] (x1) at (0.5,2.5) {};
\draw[->] (0.5,3)--(x1.north) {};

\node[] at (2.1,3.0) {Input from path $X$};
\node[] at (2.1,2.3) {\ldots};
\node[redondo=1cm,label={right:$x_n$}] (xn) at (3.65,2.5) {};
\draw[->] (3.65,3)--(xn.north) {};

\node[redondo=0cm,label={[yshift=0.2ex]below:\textbf{i}$_1$}] (i1) at (0.6,1.65) {\begin{tabular}{c} \\[-0.3cm] \end{tabular}};
%\draw[] (1.75,1) circle (0.08cm);
\%draw[] (1.45,1) circle (0.08cm);
\node[redondo=0cm,label={below:\textbf{m}$_0$}] (m0) at (-0.75,1.0) 
	{\begin{tabular}{c} \\[-0.3cm] \end{tabular}};
%\draw[] (-0.5,1.15) circle (0.08cm);
%\draw[] (-0.5,0.85) circle (0.08cm);

\node[redondo=0cm,label={[xshift=0.5ex,yshift=0.5ex]below:\textbf{f}$_1$}] (f1) at (0,0.7)
	{\begin{tabular}{c} \\[-0.3cm] \end{tabular}};

\node at (1.12,-0.05) {LSTM cell};

\node[redondo=0cm,label={below:\textbf{m}$_1$}] (m1) at (1.3,1.0) 
	{\begin{tabular}{c} \\[-0.3cm] \end{tabular}};
%\draw[] (4,1.15) circle (0.08cm);
%\draw[] (4,0.85) circle (0.08cm);
\node[redondo=0cm,label={below:\textbf{o}$_1$}] (o1) at (2.25,0.7) 
	{\begin{tabular}{c} \\[-0.3cm] \end{tabular}};
%\draw[] (4.5,-0.15) circle (0.08cm);
%\draw[] (4.5,-0.45) circle (0.08cm);
	
\node[redondo=0cm,label={below:\textbf{e}$_1$}] (e1) at (2.75,0.7) 
	{\begin{tabular}{c} \\[-0.3cm] \end{tabular}};
%\draw[] (5.5,-0.15) circle (0.08cm);
%\draw[] (5.5,-0.45) circle (0.08cm);	
\node[] at (3.3,0.8) {\ldots};

\draw[very thick] (m0.east)--(f1.west);

\draw[very thick] (m0.east)--(i1.west);

\draw[very thick] (x1.south)--(i1.north);

\draw[very thick] (x1.south)--(f1.north);

\draw[very thick] (i1.east)--(m1.west);
\draw[very thick] (f1.east)--(m1.west);

\draw[very thick] (x1.south)--(o1.west);

\draw[very thick] (m1.east)--(o1.west);

\draw[very thick] (o1.east)--(e1.west);

\node[] (on) at (3.5,0.7) {};
\node[] (in) at (3.65,1.5) {};
\node[] (mn1) at (3.4,1.25) {};
\node[] (mn2) at (3.6,1.0) {};

\node[redondo=0cm,label={below:\textbf{e}$_n$}] (en) at (4,0.7) 
	{\begin{tabular}{c} \\[-0.3cm] \end{tabular}};
\draw[-,dashed, very thick] (m1.east)--(mn1.west);
\draw[-,dashed, very thick] (xn.south)--(in.north);
\node[] at (1.25,-1.1) {\begin{tabular}{c} Vector of binary indicator \\ features   $B$ \end{tabular}};
\node[] at (3.25,-1.8) {(2)};
\node[redondo=0cm] (B) at (3.75,-1.3) 
	{\begin{tabular}{c} \\ \\ \\[-0.3em] \end{tabular}};
\draw[->] (2.25,-1.3)--(B.west);

\node[] at (4.5,-1.4) {(3)};
\node[redondo=0cm,label={above:\textbf{h}}] (h) at (4.75,-0.5) 
	{\begin{tabular}{c} \\ \\ \end{tabular}};
\draw[very thick] (en.east)--(h.west);
\draw[very thick] (B.east)--(h.west);
\node[] at (5.35,-1.2) {(4)};
\node[redondo=0cm,label={above:\textbf{s}}] (s) at (5.5,-0.5) 
	{\begin{tabular}{c} \\ \end{tabular}};
%\draw[] (11,-1.15) circle (0.08cm);
%\draw[] (11,-0.85) circle (0.08cm);	
\draw[->] (s.east)--(6,-0.5);

\node[] at (6.25,0) {\begin{tabular}{c}class \\ label \\ $c$ \end{tabular}};
\draw[very thick] (h.east)--(s.west);

\end{tikzpicture}
}}
\caption{Neural model for joint learning of path embeddings and higher-order features: The path sequence $x_1 \ldots x_n$ is fed into a LSTM layer, a hidden layer \textbf{h} combines the final embedding \textbf{e}$_n$ and binary input features $B$, and an output layer \textbf{s} assigns the highest probable class label $c$.}
\label{fig:network}
\end{figure}

\subsection{Joint Embedding and Feature Learning}

Our SRL model consists of four components depicted in Figure~\ref{fig:network}: 
(1) an LSTM component takes lexicalized dependency paths as input, (2) an additional input layer takes binary features as input, (3) a hidden layer combines dependency path embeddings and binary features using rectified linear units, and (4) a softmax classification layer produces output based on the hidden layer state as input. We therefore learn path embeddings jointly with feature detectors based on traditional, binary indicator features.

Given a dependency path $X$, with steps $x_k\in\{x_1,...,x_n\}$, and a
set of binary features $B$ as input, we use the LSTM formalization
from equations~(1--5) to compute the embedding $\textbf{e}_n$ at time
step $n$ and formalize the state of the hidden layer~\textbf{h} and
softmax output \textbf{s}$_c$ for each class category $c$ as follows:

\vspace{-0.5em}
\begin{gather}
%\textbf{e}_n = \textbf{o}_n \odot \sigma (\textbf{m}_n) \\
\textbf{h} = \text{max}(0, \textbf{W}^{\text{Bh}}B + \textbf{W}^{\text{eh}} \textbf{e}_n + \textbf{b}^\text{h}) \\
\textbf{s}_c = \frac{ \textbf{W}^{\text{es}}_c\textbf{e}_n + \textbf{W}^{\text{hs}}_c\textbf{h} + \textbf{b}^\text{s}_c}{\Sigma_i (\textbf{W}^{\text{es}}_i\textbf{e}_n + \textbf{W}^{\text{hs}}_i \textbf{h} + \textbf{b}^\text{s}_i) }
\end{gather}

\section{System Architecture}
\label{sec:model}

\begin{figure}[t]
\fbox{\vbox{
\begin{tikzpicture}[
  %->,
  >=stealth'
  ]
  
  \node at (4,13) {};
  \node at (4,12.1) {He had trouble \underline{raising} funds.};
  \node[] at (0,12.1) {\begin{tabular}{@{}l} \textsc{Predicate} \\ \textsc{\small{Identification}} \end{tabular}};

  \draw (-0.75,11.2)--(0.25,11.2);
  \draw[loosely dotted] (1,11.2)--(6.3,11.2);
  
  \node[] at (0.1,10.3) {\begin{tabular}{@{}l} \textsc{Predicate} \\ \textsc{\small{Disambiguation}} \end{tabular}};
  \node[draw, rounded corners] (raise) at (4,10) {raise.01};
  \draw[->] (4.7,11.8)-- node[label={[xshift=-0.1cm,yshift=-0.2cm]left:sense}] {} (raise.north);
  
  \draw (-0.75,9.4)--(0.25,9.4);
  \draw[loosely dotted] (1,9.4)--(6.3,9.4);  
  
  \node[] at (0,8.5) {\begin{tabular}{@{}l} \textsc{Argument} \\ \textsc{\small{Identification}} \end{tabular}};

  \node[] (arg2) at (5.75,8.1) {ARG?};
  \node[] (arg1) at (2.1,8.1) {ARG?};
  \draw[->] (2.1,11.8)--(2.1,8.8)-- node[label={right:2nd best arg}] {} (arg1.north);
 \draw[->] (5.75,11.8)--(5.75,9.8)-- node[label={left:1st best arg}] {} (arg2.north);
  
  \draw (-0.75,7.6)--(0.25,7.6);
  \draw[loosely dotted] (1,7.6)--(6.3,7.6);  
 
  \node[] at (0.02,6.7) {\begin{tabular}{@{}l} \textsc{Argument} \\ \textsc{\small{Classification}} \end{tabular}};
  \node[draw,rounded corners] (arg1A0) at (1.6,6.3) {A0};
  \node[draw,rounded corners] (arg2A1) at (3.9,6.3) {A1};
  \node[draw,rounded corners] (arg2A0) at (5.85,6.2) {A0};
  \draw[->] (arg1.south)--node[label={[yshift=-0.35cm,xshift=-0.1cm]right:best label}] {} (arg1A0.north);
  \draw[->] (arg2.south)--node[label={[yshift=-0.15cm,xshift=-0.8cm]above:best label}] {} (arg2A1.north);  
  \draw[->] (arg2.south)--node[label={[yshift=0.0cm]left:2nd}] {} (arg2A0.north); 
  \node[] at (5.4,6.75) {best};
    
  \draw (-0.75,5.8)--(0.25,5.8);
  \draw[loosely dotted] (1,5.8)--(6.3,5.8);

  \node[] at (-0.1,4.9) {\begin{tabular}{@{}l} \\[-0.6em] \textsc{Reranker} \\[-0.6em] ~ \end{tabular}};
  \draw (-0.75,4)--(0.25,4);  
  \draw[loosely dotted] (1,4)--(6.3,4);  
  \node[draw, rounded corners] (rr1A0) at (1.25,4.85) {he};
  \node[draw, rounded corners] (rr1A1) at (2.15,4.85) {funds};
  \node[draw, rounded corners] (rr2A1) at (3.9,4.85) {funds};
  \node[draw, rounded corners] (rr2A0) at (5.2,4.85) {funds};

%  \node[redondo=1.2cm,gray,fill=white] (rr1) at (1.6,5.1) {\begin{tabular}{c} \end{tabular}};
  \node[draw, rounded corners,fill=white] (rr1) at (1.6,5.3) {raise.01};
%  \node[redondo=1.2cm,gray,fill=white] (rr2) at (3.5,5.1) {\begin{tabular}{c} \end{tabular}};
  \node[draw, rounded corners,fill=white] (rr2) at (3.5,5.3) {raise.01};
%  \node[redondo=1.2cm,gray,fill=white] (rr3) at (5.6,5.1) {\begin{tabular}{c} \end{tabular}};
  \node[draw, rounded corners,fill=white] (rr3) at (5.6,5.3) {raise.01};
  \draw[decorate,decoration={brace,amplitude=3pt,mirror}] 
    (0.95,4.55)--node[label={below:A0}] {} (1.5,4.55); 
  \draw[decorate,decoration={brace,amplitude=3pt,mirror}] 
    (1.85,4.545)--node[label={below:A1}] {} (2.45,4.545);   
  
  \draw[dashed] (rr1.south)--(1.8,3.3);
 \draw[dashed,->] (2,3.3)-|(1.6,2.9);
 \draw[dashed,->] (2,3.3)-|node[label={[xshift=-1.8cm]above:best overall scoring structure}] {} (5.9,2.9);
  \node[] (score2) at (3.25,4.25) {score};  
  \draw[dotted,->] ([xshift=-0.25cm]rr2.south)--(score2.north);
  \node[] (score3) at (5.85,4.25) {score};  
  \draw[dotted,->] ([xshift=+0.25cm]rr3.south)--(score3.north);
  
  \draw[->,thick] (arg1A0.south)--(rr1.north);
  \draw[->,thick] (arg2A1.south)--([xshift=0.4cm]rr1.north);  
  \draw[->,thick] (arg2A1.south)--([xshift=0.4cm]rr2.north);  
  \draw[->,thick] (arg2A0.south)--(rr3.north);  
    
  \node[] at (-0.25,3.1) {\textsc{Output}};
  \node[] at (-0.25,2.4) {};
  \node at (3.6,2.7) {He$_\text{A0}$ had trouble \textbf{raising} funds$_\text{A1}$.};
  
\end{tikzpicture}
}}
\caption{Pipeline architecture of our SRL system.}
\label{fig:system}
\end{figure}
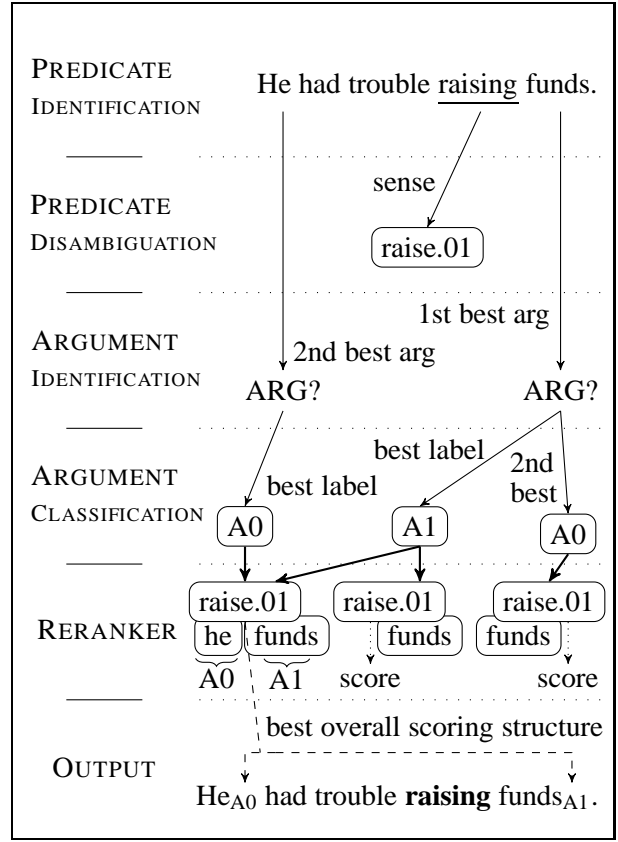

The overall architecture of our SRL system closely follows that of previous work \cite{toutanova08,bjoerkelund09} and is depicted in Figure~\ref{fig:system}. 
We use a pipeline that consists of the following steps: predicate identification and disambiguation, argument identification, argument classification, and re-ranking. The neural-network components introduced in Section~\ref{sec:depemb} are used in the last three steps. The following sub-sections describe all components in more detail.

\subsection{Predicate Identification and Disambiguation}

Given a syntactically analyzed sentence, the first two steps in an
end-to-end SRL system are to identify and disambiguate the semantic
predicates in the sentence. Here, we focus on verbal and nominal
predicates but note that other syntactic categories have also been
construed as predicates in the NLP literature (e.g.,\ prepositions;
\newcite{srikumar13}). For both identification and disambiguation
steps, we apply the same logistic regression classifiers used in the
SRL components of mate-tools \cite{bjoerkelund10}. The classifiers for
both tasks make use of a range of lexico-syntactic indicator features,
including predicate word form, its predicted part-of-speech tag as
well as dependency relations to all syntactic children.

\subsection{Argument Identification and Classification}

Given a sentence and a set of sense-disambiguated predicates in it,
the next two steps of our SRL system are to identify all arguments of
each predicate and to assign suitable role labels to them. For both
steps, we train several LSTM-based neural network models as described
in Section~\ref{sec:depemb}. In particular, we train separate networks
for nominal and verbal predicates and for identification and
classification. Following the findings of earlier work \cite{xue04},
we assume that different feature sets are relevant for the respective
tasks and hence different embedding representations should be
learned. As binary input features, we use the following sets from the
SRL literature \cite{bjoerkelund10}.

\begin{table*}
\begin{center}
\begin{tabular}{ccccccc}
\toprule
Argument labeling step &  forget gate & memory$\rightarrow$gates & $|e|$& $|h|$ & alpha & dropout rate \\
\midrule
Identification (verb) & $-$ & $+$ & 25 & ~90 & 0.0006 & 0.42 \\
Identification (noun) & $-$ & $+$ & 16 & 125 & 0.0009 & 0.25 \\
Classification (verb) & $+$ & $-$ & ~5 & 300 & 0.0155 & 0.50 \\
Classification (noun) & $-$ & $-$ & 88 & 500 & 0.0055 & 0.46 \\
\bottomrule
\end{tabular}
\end{center}
\caption{Hyperparameters selected for best models and training procedures}
\label{tbl:hyperparams}
\end{table*}

\paragraph{Lexico-syntactic features} Word form and word category of the predicate and candidate argument; dependency relations from predicate and argument to their respective syntactic heads; full dependency path sequence from predicate to argument.

\paragraph{Local context features} Word forms and word categories of the candidate argument's and predicate's syntactic siblings and children words.

\paragraph{Other features} Relative position of the candidate argument with respect to the predicate (left, self, right); sequence of part-of-speech tags of all words between the predicate and the argument.

\subsection{Reranker}
\label{subsec:reranker}
As all argument identification (and classification) decisions are independent of one another, we apply as the last step of our pipeline a global reranker. Given a predicate $p$, the reranker takes as input the $n$ best sets of identified arguments as well as their $n$ best label assignments and predicts the best overall argument structure. We implement the reranker as a logistic regression classifier, with hidden and embedding layer states of identified arguments as features, offset by the argument label, and a binary label as output (1:\ best predicted structure, 0:\ any other structure). At test time, we select the structure with the highest overall score, which we compute as the geometric mean of the global regression and all argument-specific scores.

\section{Experiments}
\label{sec:exp}

In this section, we demonstrate the usefulness of dependency path
embeddings for semantic role labeling. Our hypotheses are that (1)
modeling dependency paths as sequences will lead to better
representations for the SRL task, thus increasing labeling precision
overall, and that (2) embeddings will address the problem of data
sparsity, leading to higher recall. To test both hypotheses, we
experiment on the in-domain and out-of-domain test sets provided in
the CoNLL-2009 shared task \cite{hajic09} and compare results of our
system, henceforth PathLSTM, with systems that do not involve path
embeddings.  We compute precision, recall and F$_1$-score using the
official CoNLL-2009 scorer.\footnote{Some recently proposed SRL models
  are only evaluated on the CoNLL 2005 and 2012 data sets, which lack
  nominal predicates or dependency annotations. We do not list any
  results from those models here.}  The code is available at \url{
  https://github.com/microth/PathLSTM}.

\paragraph{Model selection} We train argument identification and classification models using the XLBP toolkit for neural networks \cite{monner12}. The hyperparameters for each step were selected based on the CoNLL 2009 development set. For direct comparison with previous work, we use the same preprocessing models and predicate-specific SRL components as provided with mate-tools \cite{bohnet10,bjoerkelund10}. The types and ranges of hyperparameters considered are as follows: learning rate $\alpha \in [0.00006,0.3]$,  dropout rate $d \in [0.0,0.5]$, and hidden layer sizes $|e| \in [0,100]$, $|h| \in [0,500]$. In addition, we experimented with different gating mechanisms (with/without forget gate) and memory access settings (with/without connections between all gates and the memory layer, cf.\ Section~\ref{sec:depemb}). The best parameters were chosen using the Spearmint hyperparameter optimization toolkit \cite{snoek12}, applied for approx.\ 200 iterations, and are summarized in Table~\ref{tbl:hyperparams}.

\paragraph{Results} The results of our in- and out-of-domain
experiments are summarized in Tables~\ref{tbl:results}
and~\ref{tbl:oodresults}, respectively.  We present results for
different system configurations: `local' systems make classification
decisions independently, whereas `global' systems include a reranker
or other global inference mechanisms; `single' refers to one model and
`ensemble' refers to combinations of multiple models.

In the in-domain setting, our PathLSTM model achieves 87.7\% (single) and 87.9\% (ensemble) F$_1$-score, outperforming previously published best results by 0.4 and 0.2 percentage points, respectively. At a F$_1$-score of 86.7\%, our local model (using no reranker) reaches the same performance as state-of-the-art local models. Note that differences in results between systems might originate from the application of different preprocessing techniques as each system comes with its own syntactic components. For direct comparison, we evaluate against mate-tools, which use the same preprocessing techniques as PathLSTM. In comparison, we see improvements of $+$0.8--1.0 percentage points absolute in F$_1$-score.

\begin{savenotes}
\begin{table}[t]
\begin{tabular}{@{ }l@{~~~}ccc@{ }}
\toprule
\multicolumn{1}{c}{System \footnotesize{(local, single)}} & P & R & F$_1$ \\
\midrule
\newcite{bjoerkelund10} & 87.1 & 84.5 & 85.8\\
\newcite{lei15} & $-$ & $-$ & 86.6\\
\newcite{fitzgerald15} & $-$ & $-$ & \textbf{86.7} \\
PathLSTM w/o reranker & \textbf{88.1} & \textbf{85.3} & \textbf{86.7} \\
\midrule
\midrule
\multicolumn{1}{c}{System \footnotesize{(global, single)}} & P & R & F$_1$ \\
\midrule
\newcite{bjoerkelund10} & 88.6 & 85.2 & 86.9\\
\newcite{rothwoodsend14}\footnote{Results are taken from \newcite{lei15}.} & $-$ & $-$ & 86.3 \\
\newcite{fitzgerald15} & $-$ & $-$ & 87.3 \\
PathLSTM & \textbf{90.0} & \textbf{85.5} & \textbf{87.7} \\
\midrule
\midrule
\multicolumn{1}{c}{System \footnotesize{(global, ensemble)}} & P & R & F$_1$ \\
\midrule
FitzGerald et al. \hfill 10 models & $-$ & $-$ & 87.7 \\
PathLSTM\hfill 3 models & \textbf{90.3} & \textbf{85.7} & \textbf{87.9} \\
\bottomrule
\end{tabular}
\caption{Results on the CoNLL-2009 in-domain test set. All numbers are in percent.}
\label{tbl:results}
\end{table}
\end{savenotes}

\begin{table}[t]
\begin{tabular}{@{ }lccc@{ }}
\toprule
PathLSTM & P (\%) & R (\%) & F$_1$ (\%) \\
\midrule
w/o path embeddings & 65.7 & 87.3 & 75.0 \\
w/o binary features & 73.2 & 33.3 & 45.8 \\
\bottomrule
\end{tabular}
\caption{Ablation tests in the in-domain setting.}
\label{tbl:ablation}
\end{table}

\begin{table}[t]
\begin{tabular}{@{ }l@{~~~}ccc@{ }}
\toprule
\multicolumn{1}{c}{System \footnotesize{(local, single)}} & P & R & F$_1$ \\
\midrule
\newcite{bjoerkelund10} & 75.7 & 72.2 & 73.9 \\
\newcite{lei15} & $-$ & $-$ & \textbf{75.6}\\
\newcite{fitzgerald15} & $-$ & $-$ & 75.2\\
PathLSTM w/o reranker & \textbf{76.9} & \textbf{73.8} & 75.3 \\
\midrule
\midrule
\multicolumn{1}{c}{System \footnotesize{(global, single)}} & P & R & F$_1$ \\
\midrule
\newcite{bjoerkelund10} & 77.9 & 73.6 & 75.7\\
\newcite{rothwoodsend14}$^\text{3}$ & $-$ & $-$ & 75.9 \\
\newcite{fitzgerald15} & $-$ & $-$ & 75.2 \\
PathLSTM & \textbf{78.6} & \textbf{73.8} & \textbf{76.1} \\
\midrule
\midrule
\multicolumn{1}{c}{System \footnotesize{(global, ensemble)}} & P & R & F$_1$ \\
\midrule
FitzGerald et al. \hfill 10 models & $-$ & $-$ & 75.5 \\
PathLSTM \hfill 3 models & {79.7} & 73.6 & \textbf{76.5} \\
\bottomrule
\end{tabular}
\caption{Results on the CoNLL-2009 out-of-domain test set. All numbers are in percent.}
\label{tbl:oodresults}
\end{table}

In the out-of-domain setting, our system achieves new state-of-the-art results of 76.1\% (single) and 76.5\% (ensemble) F$_1$-score, outperforming the previous best system by \newcite{rothwoodsend14} by 0.2 and 0.6 absolute points, respectively. In comparison to mate-tools, we observe absolute improvements in F$_1$-score of $+$0.4--0.8\%.

\paragraph{Discussion} 
To determine the sources of individual improvements,
we test PathLSTM models without specific feature types and directly compare PathLSTM and mate-tools, both of which use the same preprocessing methods. 
%We analyse their output qualitatively and perform quantitative comparisons considering three aspects: predicate word category, role label, and sentence length.
Table~\ref{tbl:ablation} presents in-domain test results for our
system when specific feature types are omitted. The overall low
results indicate that a combination of dependency path embeddings and
binary features is required to identify and label arguments  with
high precision.

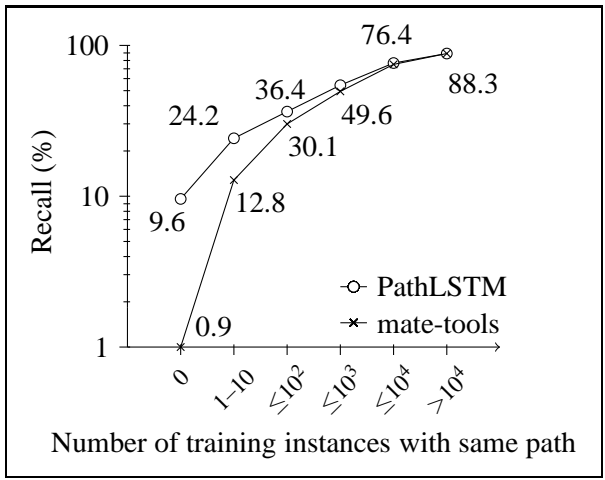
\begin{figure}[t]
\vbox{\fbox{
\begin{tikzpicture}[y=.2cm, x=.7cm]
 	%axis
	\draw[->] (0,0) -- coordinate (x axis mid) (7,0);
    	\draw (0,2.6) -- coordinate (y axis mid) (0,20);
    	\draw (0,0)--(0,2.6);

    	\foreach \x in {1,...,6}
     		\draw (\x,1pt) -- (\x,-3pt)
			node[anchor=north] {};
		\node[rotate=45] at (1,-2.0) {\footnotesize{0}};
		\node[rotate=45] at (2,-2.8) {\footnotesize{1--10}};
		\node[rotate=45] at (3,-2.8) {\footnotesize{$\leq$10$^2$}};
		\node[rotate=45] at (4,-2.8) {\footnotesize{$\leq$10$^3$}};
		\node[rotate=45] at (5,-2.8) {\footnotesize{$\leq$10$^4$}};
		\node[rotate=45] at (6,-2.8) {\footnotesize{$>$10$^4$}};								
	   	\foreach \y in {100}
     		\draw (1pt,\y*0.2) -- (-3pt,\y*0.2) 
     			node[anchor=east] {\y};
	   	\foreach \y in {0.3,0.48,0.6,0.7,0.78,0.845,0.9,0.95,
	   	                1.3,1.48,1.6,1.7,1.78,1.845,1.9,1.95}
     		\draw (1pt,\y*10) -- (-1.5pt,\y*10) 
     			node[anchor=east] {};

	\draw (1pt,0) -- (-3pt,0*0.2) 
		node[anchor=east] {1};
	\draw (1pt,50*0.2) -- (-3pt,50*0.2) 
		node[anchor=east] {10};

	%labels      
	\node[below=1cm] at (x axis mid) {Number of training instances with same path};
	\node[rotate=90, above=0.8cm] at ([xshift=-0.8cm,yshift=-1cm]y axis mid) {Recall (\%)};

	%plots
	\draw plot[mark=*, mark options={fill=white}] 
		file {NN+RR_R.data};
	\draw plot[mark=x, mark options={fill=white} ] 
		file {mate_R.data};
 
	%legend
	\begin{scope}[shift={(4,4)}] 
	\draw (0,0) -- 
		plot[mark=*, mark options={fill=white}] (0.25,0) -- (0.5,0) 
		node[right]{PathLSTM};
    \draw[yshift=-\baselineskip] (0,0) -- 
		plot[mark=x, mark options={fill=white}] (0.25,0) -- (0.5,0)
		node[right]{mate-tools};
	\end{scope}
	
   \node[] at (1.625,1.375) {0.9};
   \node[] at (0.75,98.2*0.1-1.5) {9.6};
   \node[] at (1.25,138.4*0.1+1.5) {24.2};
   \node[] at (2.5,110.7*0.1-1.5) {12.8};

   \node[] at (2.875,156.1*0.1+1.5) {36.4};
   \node[] at (3.5,147.9*0.1-1.5) {30.1};
   
   \node[] at (4.5,169.5*0.1-1.5) {49.6};
   \node[] at (4.875,188.3*0.1+1.875) {76.4};
   \node[] at (6.5,194.6*0.1-2) {88.3};
	
\end{tikzpicture}
}}
\caption{Results on in-domain test instances, grouped by the number of training instances that have an identical (unlexicalized) dependency path.}
%Note that a logarithmic scale is used for both axes.}
\label{fig:sparseresults}
\end{figure}

Figure~\ref{fig:sparseresults} shows the effect of dependency path
embeddings at mitigating sparsity: if the path between a predicate and
its argument has not been observed at training time or only
infrequently, conventional methods will often fail to assign a
role. This is represented by the recall curve of mate-tools, which
converges to zero for arguments with unseen paths. The higher recall
curve for PathLSTM demonstrates that path embeddings can alleviate
this problem to some extent. For unseen paths, we observe that
PathLSTM improves over mate-tools by an order of magnitude, from 0.9\%
to 9.6\%. The highest absolute gain, from 12.8\% to 24.2\% recall, can
be observed for dependency paths that occurred between 1 and~10 times
during training.

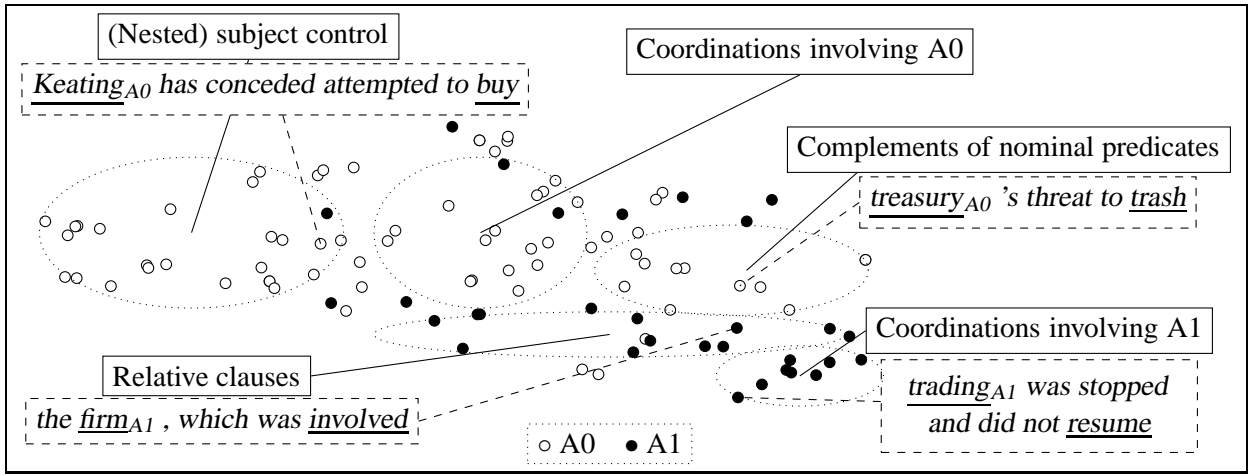
\begin{figure*}[t]
\fbox{
\begin{tikzpicture}
	\draw plot[mark=*, only marks,mark options={fill=white}] file {objcontrol.data};
	\draw plot[mark=*, only marks] file {A1coord.data};	
	\draw plot[mark=*, only marks] file {relclause.data};	
	\draw[dotted] (-4.0,0.4) ellipse (2.0cm and 1cm);
	\draw[dotted] (4,-1.5) ellipse (1.1cm and 0.4cm);
    \draw[dotted] (1.5,-0.95) ellipse (3.1cm and 0.3cm);

    \node[draw] (obj) at (-3.25,3) {(Nested) subject control};
    \draw[-] (obj.south)--(-4,0.4);
    \node[draw] (coord1) at (7.15,-0.9) {Coordinations involving A1};
    \draw[-] (coord1.west)--(4,-1.5);
    \node[draw] (relclause) at (-3.8,-1.5) {Relative clauses};
    \draw[-] (relclause.east)--(1.5,-0.95);    

	\draw plot[mark=*, only marks,mark options={fill=white}] file {A0coord.data};
    \draw plot[mark=*, only marks,mark options={fill=white}] file {nompreds.data};	
    \draw[dotted] (-0.2,0.4) ellipse (1.4cm and 1.0cm);
    \draw[dotted] (3.1,-0.1) ellipse (1.8cm and 0.6cm);

	\node[draw] (coord0) at (4,2.8) {Coordinations involving A0};
	\draw[-] (coord0.south)--(-0.2,0.4);
	\node[draw] (nompreds) at (6.75,1.5) {Complements of nominal predicates};
	\draw[-] ([xshift=-2cm]nompreds.south)--(3.3,-0.1);
	        
    \node[dashed,draw] (nompredex) at (7,0.8) {\textsl{\underline{treasury}$_\text{A0}$ 's threat to \underline{trash}}};
    \draw[dashed,-] (nompredex.west)--(3.2124,-1.5223/5);
    
    \node[dashed,draw] (relex) at (-3.6,-2.1) {\textsl{the \underline{firm}$_\text{A1}$ , which was \underline{involved}}};
    \draw[dashed,-] (relex.east)--(3.1718,-4.3220/5);
    
    \node[dashed,draw,fill=white] (controlex) at (-2.9,2.3) {\textsl{\underline{Keating}$_\text{A0}$ has conceded attempted to \underline{buy}}};
    \draw[dashed,-] (controlex)--(-2.3026,1.2620/5);

    \node[dashed,draw] (coordex) at (7.15,-1.9) {\begin{tabular}{c} \textsl{\underline{trading}$_\text{A1}$ was stopped} \\ \textsl{and did not \underline{resume}}\end{tabular}};
    \draw[dashed,-] (coordex)--(    3.1867,-8.9270/5);
    
    \node[draw,dotted] at (1.5,-2.4) {$\circ$ A0\quad $\bullet$ A1};

\end{tikzpicture}
}
\caption{Dots correspond to the path representation of a
  predicate-argument instance in 2D space. White/black color indicates
  A0/A1 gold argument labels. Dotted ellipses denote instances
  exhibiting related syntactic phenomena (see rectangles for a
  description and dotted rectangles for linguistic examples).  Example
  phrases show actual output produced by PathLSTM (underlined).}
\label{tbl:qualeval}
%%%\vspace{-0.5em}
\end{figure*}

\begin{figure}[t]
\vbox{\fbox{
\begin{tikzpicture}[y=.2cm, x=.7cm]
 	%axis
	\draw[->] (0,0) -- coordinate (x axis mid) (7,0);
    	\draw (0,2.6) -- coordinate (y axis mid) (0,20);
    	\draw (0,0)--(0,2.6);

    	%ticks
    	\foreach \x in {1,...,6}
     		\draw (\x,1pt) -- (\x,-3pt)
			node[anchor=north] {};
		\node[rotate=45] at (1,-2.2) {\footnotesize{1--10}};
		\node[rotate=45] at (2,-2.2) {\footnotesize{11--15}};
		\node[rotate=45] at (3,-2.2) {\footnotesize{16--20}};
		\node[rotate=45] at (4,-2.2) {\footnotesize{21--25}};
		\node[rotate=45] at (5,-2.2) {\footnotesize{26--30}};
		\node[rotate=45] at (6,-2.2) {\footnotesize{31--}};								
    	\foreach \y in {85,86,...,90}
     		\draw (1pt,\y*3-250) -- (-3pt,\y*3-250) 
     			node[anchor=east] {\y}; 
	%labels      
	\node[below=1.0cm] at (x axis mid) {Number of words};
	\node[rotate=90, above=0.8cm] at ([xshift=-0.8cm,yshift=-0.5cm]y axis mid) {F$_1$-score (\%)};

	%plots
	\draw plot[mark=*, mark options={fill=white}] 
		file {NN+RR_F.data};
	\draw plot[mark=x, mark options={fill=white} ] 
		file {mate_F.data};
 
	%legend
	\begin{scope}[shift={(4,4)}] 
	\draw (0,0) -- 
		plot[mark=*, mark options={fill=white}] (0.25,0) -- (0.5,0) 
		node[right]{PathLSTM};
    \draw[yshift=-\baselineskip] (0,0) -- 
		plot[mark=x, mark options={fill=white}] (0.25,0) -- (0.5,0)
		node[right]{mate-tools};
	\end{scope}
	
	%labels
	\node[] at (1,89.3*3-250-1.5) {89.3};
	\node[] at (2,90.1*3-250+1.25) {90.1 (\underline{$+$0.8})};
	
	\node[] at (6,86.1*3-250-1.5) {86.1};
	\node[] at (7,87.1*3-250+1.25) {87.1 (\underline{$+$1.0})};

	\node[] at (3.5,87.5*3-250-1.5) {87.5};
	\node[] at (5,87.9*3-250+1.75) {87.9 (\underline{$+$0.4})};

\end{tikzpicture}
}}
\caption{Results by sentence length. Improvements over mate-tools shown
  in parentheses.} 
\label{fig:lengthresults}
\end{figure}
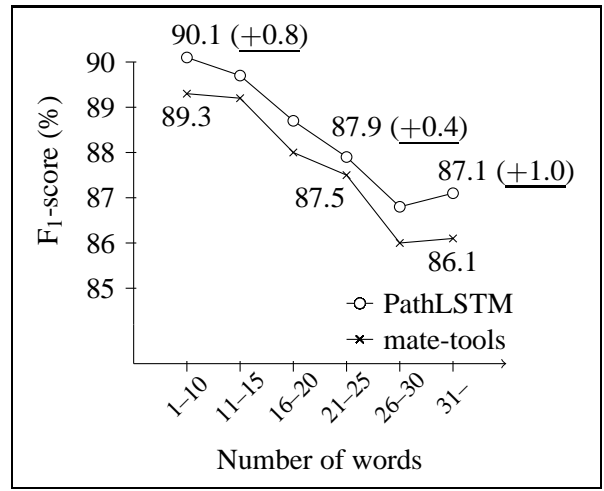

Figure~\ref{fig:lengthresults} plots role labeling performance for
sentences with varying number of words. There are two categories of
sentences in which the improvements of PathLSTM are most noticeable:
Firstly, it better handles short sentences that contain expletives
and/or nominal predicates ($+0.8\%$ absolute in F$_1$-score). This is
probably due to the fact that our learned dependency path
representations are lexicalized, making it possible to model argument
structures of different nominals and distinguishing between expletive
occurrences of `it' and other subjects. Secondly, it improves
performance on longer sentences (up to $+1.0\%$ absolute in
F$_1$-score). This is mainly due to the handling of dependency paths
that involve complex structures, such as coordinations, control verbs
and nominal predicates.

We collect instances of different syntactic phenomena from the
development set and plot the learned dependency path representations
in the embedding space (see Figure~\ref{tbl:qualeval}). We obtain a
projection onto two dimensions using t-SNE \cite{vandermaaten08}.
Interestingly, we can see that different syntactic configurations are
clustered together in different parts of the space and that most
instances of the PropBank roles A0 and A1 are separated. Example
phrases in the figure highlight predicate-argument pairs that are
correctly labeled by PathLSTM but not by mate-tools. Path embeddings
are essential for handling these cases as indicator features do not
generalize well enough.

Finally, Table~\ref{tbl:poslabelresults} shows results for nominal and
verbal predicates as well as for different (gold) role labels. In
comparison to mate-tools, we can see that PathLSTM improves precision
for all argument types of nominal predicates. For verbal predicates,
improvements can be observed in terms of recall of proto-agent (A0)
and proto-patient (A1) roles, with slight gains in precision for the
A2 role. Overall, PathLSTM does slightly worse with respect to
modifier roles, which it labels with higher precision but at the cost
of recall.

\section{Path Embeddings in other Languages}
\label{sec:multiling}

In this section, we report results from additional experiments on Chinese, German and Spanish data. The underlying question is to which extent the improvements of our SRL system for English also generalize to other languages. To answer this question, we train and test separate SRL models for each language, using the system architecture and hyperparameters discussed in Sections~\ref{sec:model} and~\ref{sec:exp}, respectively. 

We train our models on data from the \mbox{CoNLL-2009} shared task,
relying on the same features as one of the participating systems
\cite{bjoerkelund09}, and evaluate with the official scorer. For
direct comparison, we rely on the (automatic) syntactic preprocessing
information provided with the CoNLL test data and compare our results
with the best two systems for each language that make use of the same
preprocessing information.

The results, summarized in Table~\ref{tbl:multilingresults}, indicate that PathLSTM performs better than the system by \newcite{bjoerkelund09} in all cases. For German and Chinese, PathLSTM achieves the best overall F$_1$-scores of 80.1\% and 79.4\%, respectively.

\begin{table}[t]
\begin{tabular}{@{}c@{ ~}c@{~~~~}c@{ ~}c c@{ ~}c@{}}
\toprule
Predicate POS & & & \multicolumn{3}{c}{~Improvement} \\
 \& Role Label    &   \multicolumn{2}{c}{PathLSTM~~~}      & \multicolumn{3}{c}{~over mate-tools} \\
\midrule 
	& P (\%) & R (\%) && P (\%)& R (\%) \\         
\midrule
verb / A0 & 90.8 & \textbf{89.2} && $-$0.4 & $+$1.8 \\
verb / A1 & 91.0 & \textbf{91.9} && $+$0.0 & $+$1.1 \\
verb / A2 & \textbf{84.3} & 76.9 && $+$1.5 & $+$0.0 \\
verb / AM & \textbf{82.2} & 72.4 && $+$2.9 & $-$2.0 \\
\midrule
noun / A0 & \textbf{86.9} & \textbf{78.2} && $+$0.8 & $+$3.3 \\
noun / A1 & \textbf{87.5} & \textbf{84.4} && $+$2.6 & $+$2.2 \\
noun / A2 & \textbf{82.4} & \textbf{76.8} && $+$1.0 & $+$2.1 \\
noun / AM & \textbf{79.5} & 69.2 && $+$0.9 & $-$2.8 \\
\bottomrule 

\end{tabular}
\caption{Results by word category and role label.}
\label{tbl:poslabelresults}
\end{table}

\begin{table}[t]
\begin{tabular}{lccc}
\toprule
Chinese & P & R & F$_1$ \\
\midrule
PathLSTM & \textbf{83.2} & \textbf{75.9} & \textbf{79.4} \\
\newcite{bjoerkelund09} & 82.4 & 75.1 & 78.6\\
\newcite{zhao09} 		 & 80.4 & 75.2 & 77.7 \\
\midrule
\midrule
German & P & R & F$_1$ \\
\midrule
PathLSTM 			 & 81.8 & \textbf{78.5} & \textbf{80.1} \\
\newcite{bjoerkelund09} & 81.2 & 78.3 & 79.7\\
\newcite{che09} 		 & \textbf{82.1} & 75.4 & 78.6 \\
\midrule
\midrule
Spanish & P & R & F$_1$ \\
\midrule
\newcite{zhao09} 		 & 83.1 & \textbf{78.0} & \textbf{80.5} \\
%\newcite{zhaohai09}	 & 82.7 & \textbf{78.1} & 80.3 \\
PathLSTM			 & \textbf{83.2} & 77.4 & 80.2 \\
\newcite{bjoerkelund09} & 78.9 & 74.3 & 76.5 \\
\bottomrule
\end{tabular}
\caption{Results (in percentage) on the CoNLL-2009 test sets for Chinese, German and Spanish.}
\label{tbl:multilingresults}
%\vspace{-0.6em}
\end{table}

\section{Related Work}
\label{sec:relwork}

\paragraph{Neural Networks for SRL} 
\newcite{collobert11} pioneered neural networks for the task of semantic role labeling. They developed a feed-forward network that uses a convolution function over windows of words to assign SRL labels. Apart from constituency boundaries, their system does not make use of any syntactic information. \newcite{foland15} extended their model and showcased significant improvements when including binary indicator features for dependency paths. Similar features were used by \newcite{fitzgerald15}, who include role labeling predictions by neural networks as factors in a global model.

These approaches all make use of binary features derived from
syntactic parses either to indicate constituency boundaries or to
represent full dependency paths. An extreme alternative has been
recently proposed in \newcite{zhou15}, who model SRL decisions with a
multi-layered LSTM network that takes word sequences as input but no
syntactic parse information at all.

Our approach falls in between the two extremes: we rely on syntactic
parse information but rather than solely making using of sparse binary
features, we explicitly model dependency paths in a neural network
architecture.

\paragraph{Other SRL approaches} Within the SRL literature, recent
alternatives to neural network architectures include sigmoid belief
networks \cite{henderson13} as well as low-rank tensor models
\cite{lei15}. Whereas Lei et al.\ only make use of dependency paths as
binary indicator features, Henderson et al.\ propose a joint model for
syntactic and semantic parsing that learns and applies incremental
dependency path representations to perform SRL decisions. The latter
form of representation is closest to ours, however, we do not build
syntactic parses incrementally. Instead, we take syntactically
preprocessed text as input and focus on the SRL task only.% yang15?

Apart from more powerful models, most recent progress in SRL can be
attributed to novel features. For instance, \newcite{deschacht09} and
\newcite{huang10} use latent variables, learned with a hidden markov
model, as features for representing words and word
sequences. \newcite{zapirain13} propose different selection preference
models in order to deal with the sparseness of lexical
features. % wu15?
\newcite{rothwoodsend14} address the same problem with word embeddings
and compositions thereof. \newcite{rothlapata15} recently introduced
features that model the influence of discourse on role labeling
decisions.

Rather than coming up with completely new features, in this work we proposed to revisit some well-known features and represent them in a novel way that generalizes better.
Our proposed model is inspired both by the necessity to overcome the problems of sparse lexico-syntactic features and by the recent success of SRL models based on neural networks.

\paragraph{Dependency-based embeddings} The idea of embedding
dependency structures has previously been applied to tasks such as
relation classification and sentiment analysis. \newcite{xu15} and
\newcite{liu15} use neural networks to embed dependency paths between
entity pairs. To identify the relation that holds between two
entities, their approaches make use of pooling layers that detect
parts of a path that indicate a specific relation. In contrast, our
work aims at modeling an individual path as a complete sequence, in
which every item is of relevance.  \newcite{tai15} and \newcite{ma15}
learn embeddings of dependency structures representing full sentences,
in a sentiment classification task. In our model, embeddings are
learned jointly with other features, and as a result problems that may
result from erroneous parse trees are mitigated.

\section{Conclusions}
\label{sec:conclusions}

We introduced a neural network architecture for semantic role labeling
that jointly learns embeddings for dependency paths and feature
combinations.  Our experimental results indicate that our model
substantially increases classification performance, leading to new
state-of-the-art results.  In a qualitive analysis, we found that our
model is able to cover instances of various linguistic phenomena that
are missed by other methods.

Beyond SRL, we expect dependency path embeddings to be useful in
related tasks and downstream applications. For instance, our
representations may be of direct benefit for semantic and discourse
parsing tasks. The jointly learned feature space also makes our model
a good starting point for cross-lingual transfer methods that rely on
feature representation projection to induce new models
\cite{kozhevnikov14}.

\paragraph{Acknowledgements} We thank the three anonymous ACL referees
whose feedback helped to substantially improve the present paper.  The
support of the Deutsche Forschungsgemeinschaft (Research Fellowship RO
4848/1-1; Roth) and the European Research Council (award number
681760; Lapata) is gratefully acknowledged.

%The research presented in
%this paper was funded by a DFG Research Fellowship (RO 4848/1-1).

\bibliography{new-nodates}

\begin{thebibliography}{}

\bibitem[\protect\citename{Aziz \bgroup et al.\egroup }2011]{aziz11}
Wilker Aziz, Miguel Rios, and Lucia Specia.
\newblock 2011.
\newblock Shallow semantic trees for smt.
\newblock In {\em Proceedings of the Sixth Workshop on Statistical Machine
  Translation}, pages 316--322, Edinburgh, Scotland.

\bibitem[\protect\citename{Bj{\"o}rkelund \bgroup et al.\egroup
  }2009]{bjoerkelund09}
Anders Bj{\"o}rkelund, Love Hafdell, and Pierre Nugues.
\newblock 2009.
\newblock Multilingual semantic role labeling.
\newblock In {\em Proceedings of the Thirteenth Conference on Computational
  Natural Language Learning: Shared Task}, pages 43--48, Boulder, Colorado.

\bibitem[\protect\citename{Bj{\"o}rkelund \bgroup et al.\egroup
  }2010]{bjoerkelund10}
Anders Bj{\"o}rkelund, Bernd Bohnet, Love Hafdell, and Pierre Nugues.
\newblock 2010.
\newblock A high-performance syntactic and semantic dependency parser.
\newblock In {\em {Coling 2010: Demonstration Volume}}, pages 33--36, Beijing,
  China.

\bibitem[\protect\citename{Bohnet}2010]{bohnet10}
Bernd Bohnet.
\newblock 2010.
\newblock Top accuracy and fast dependency parsing is not a contradiction.
\newblock In {\em {Proceedings of the 23rd International Conference on
  Computational Linguistics}}, pages 89--97, Beijing, China.

\bibitem[\protect\citename{Che \bgroup et al.\egroup }2009]{che09}
Wanxiang Che, Zhenghua Li, Yongqiang Li, Yuhang Guo, Bing Qin, and Ting Liu.
\newblock 2009.
\newblock Multilingual dependency-based syntactic and semantic parsing.
\newblock In {\em Proceedings of the Thirteenth Conference on Computational
  Natural Language Learning: Shared Task}, pages 49--54, Boulder, Colorado.

\bibitem[\protect\citename{Collobert \bgroup et al.\egroup }2011]{collobert11}
Ronan Collobert, Jason Weston, L{\'e}on Bottou, Michael Karlen, Koray
  Kavukcuoglu, and Pavel Kuksa.
\newblock 2011.
\newblock Natural language processing (almost) from scratch.
\newblock {\em The Journal of Machine Learning Research}, 12:2493--2537.

\bibitem[\protect\citename{Deschacht and Moens}2009]{deschacht09}
Koen Deschacht and Marie-Francine Moens.
\newblock 2009.
\newblock Semi-supervised semantic role labeling using the {Latent Words
  Language Model}.
\newblock In {\em Proceedings of the 2009 Conference on Empirical Methods in
  Natural Language Processing}, pages 21--29, Singapore.

\bibitem[\protect\citename{FitzGerald \bgroup et al.\egroup
  }2015]{fitzgerald15}
Nicholas FitzGerald, Oscar T\"{a}ckstr\"{o}m, Kuzman Ganchev, and Dipanjan Das.
\newblock 2015.
\newblock Semantic role labeling with neural network factors.
\newblock In {\em Proceedings of the 2015 Conference on Empirical Methods in
  Natural Language Processing}, pages 960--970, Lisbon, Portugal.

\bibitem[\protect\citename{Foland and Martin}2015]{foland15}
William Foland and James Martin.
\newblock 2015.
\newblock Dependency-based semantic role labeling using convolutional neural
  networks.
\newblock In {\em Proceedings of the Fourth Joint Conference on Lexical and
  Computational Semantics}, pages 279--288, Denver, Colorado.

\bibitem[\protect\citename{Gildea and Jurafsky}2002]{gildea02}
Daniel Gildea and Daniel Jurafsky.
\newblock 2002.
\newblock Automatic labeling of semantic roles.
\newblock {\em Computational Linguistics}, 28(3):245--288.

\bibitem[\protect\citename{Haji{\v{c}} \bgroup et al.\egroup }2009]{hajic09}
Jan Haji{\v{c}}, Massimiliano Ciaramita, Richard Johansson, Daisuke Kawahara,
  Maria~Ant{\`o}nia Mart{\'\i}, Llu{\'\i}s M{\`a}rquez, Adam Meyers, Joakim
  Nivre, Sebastian Pad{\'o}, Jan {\v{S}}t{\v{e}}p{\'a}nek, et~al.
\newblock 2009.
\newblock {The CoNLL-2009 shared task: S}yntactic and semantic dependencies in
  multiple languages.
\newblock In {\em Proceedings of the Thirteenth Conference on Computational
  Natural Language Learning: Shared Task}, pages 1--18, Boulder, Colorado.

\bibitem[\protect\citename{Henderson \bgroup et al.\egroup }2013]{henderson13}
James Henderson, Paola Merlo, Ivan Titov, and Gabriele Musillo.
\newblock 2013.
\newblock Multilingual joint parsing of syntactic and semantic dependencies
  with a latent variable model.
\newblock {\em Computational Linguistics}, 39(4):949--998.

\bibitem[\protect\citename{Hochreiter and Schmidhuber}1997]{hochreiter97}
Sepp Hochreiter and J\"{u}rgen Schmidhuber.
\newblock 1997.
\newblock Long short-term memory.
\newblock {\em Neural Computation}, 9(8):1735--1780.

\bibitem[\protect\citename{Huang and Yates}2010]{huang10}
Fei Huang and Alexander Yates.
\newblock 2010.
\newblock Open-domain semantic role labeling by modeling word spans.
\newblock In {\em Proceedings of the 48th Annual Meeting of the Association for
  Computational Linguistics}, pages 968--978, Uppsala, Sweden.

\bibitem[\protect\citename{Johansson and Nugues}2008]{johansson08}
Richard Johansson and Pierre Nugues.
\newblock 2008.
\newblock The effect of syntactic representation on semantic role labeling.
\newblock In {\em Proceedings of the 22nd International Conference on
  Computational Linguistics}, pages 393--400, Manchester, United Kingdom.

\bibitem[\protect\citename{Khan \bgroup et al.\egroup }2015]{khan15}
Atif Khan, Naomie Salim, and Yogan~Jaya Kumar.
\newblock 2015.
\newblock A framework for multi-document abstractive summarization based on
  semantic role labelling.
\newblock {\em Applied Soft Computing}, 30:737--747.

\bibitem[\protect\citename{Kozhevnikov and Titov}2014]{kozhevnikov14}
Mikhail Kozhevnikov and Ivan Titov.
\newblock 2014.
\newblock Cross-lingual model transfer using feature representation projection.
\newblock In {\em Proceedings of the 52nd Annual Meeting of the Association for
  Computational Linguistics}, pages 579--585, Baltimore, Maryland.

\bibitem[\protect\citename{Lei \bgroup et al.\egroup }2015]{lei15}
Tao Lei, Yuan Zhang, Llu\'{i}s M\`{a}rquez, Alessandro Moschitti, and Regina
  Barzilay.
\newblock 2015.
\newblock High-order low-rank tensors for semantic role labeling.
\newblock In {\em Proceedings of the 2015 Conference of the North American
  Chapter of the Association for Computational Linguistics: Human Language
  Technologies}, pages 1150--1160, Denver, Colorado.

\bibitem[\protect\citename{Lewis \bgroup et al.\egroup }2015]{lewis15}
Mike Lewis, Luheng He, and Luke Zettlemoyer.
\newblock 2015.
\newblock Joint {A* CCG} parsing and semantic role labelling.
\newblock In {\em Proceedings of the 2015 Conference on Empirical Methods in
  Natural Language Processing}, pages 1444--1454, Lisbon, Portugal.

\bibitem[\protect\citename{Liu \bgroup et al.\egroup }2015]{liu15}
Yang Liu, Furu Wei, Sujian Li, Heng Ji, Ming Zhou, and Houfeng Wang.
\newblock 2015.
\newblock A dependency-based neural network for relation classification.
\newblock In {\em Proceedings of the 53rd Annual Meeting of the Association for
  Computational Linguistics and the 7th International Joint Conference on
  Natural Language Processing}, pages 285--290, Beijing, China.

\bibitem[\protect\citename{Ma \bgroup et al.\egroup }2015]{ma15}
Mingbo Ma, Liang Huang, Bowen Zhou, and Bing Xiang.
\newblock 2015.
\newblock Dependency-based convolutional neural networks for sentence
  embedding.
\newblock In {\em Proceedings of the 53rd Annual Meeting of the Association for
  Computational Linguistics and the 7th International Joint Conference on
  Natural Language Processing}, pages 174--179, Beijing, China.

\bibitem[\protect\citename{Monner and Reggia}2012]{monner12}
Derek Monner and James~A Reggia.
\newblock 2012.
\newblock A generalized {LSTM}-like training algorithm for second-order
  recurrent neural networks.
\newblock {\em Neural Networks}, 25:70--83.

\bibitem[\protect\citename{Osman \bgroup et al.\egroup }2012]{osman12}
Ahmed~Hamza Osman, Naomie Salim, Mohammed~Salem Binwahlan, Rihab Alteeb, and
  Albaraa Abuobieda.
\newblock 2012.
\newblock An improved plagiarism detection scheme based on semantic role
  labeling.
\newblock {\em Applied Soft Computing}, 12(5):1493--1502.

\bibitem[\protect\citename{Palmer \bgroup et al.\egroup }2005]{palmer05}
Martha Palmer, Daniel Gildea, and Paul Kingsbury.
\newblock 2005.
\newblock {The Proposition bank}: {A}n annotated corpus of semantic roles.
\newblock {\em Computational Linguistics}, 31(1):71--106.

\bibitem[\protect\citename{Paul and Jamal}2015]{paul15}
Merin Paul and Sangeetha Jamal.
\newblock 2015.
\newblock An improved {SRL} based plagiarism detection technique using sentence
  ranking.
\newblock {\em Procedia Computer Science}, 46:223--230.

\bibitem[\protect\citename{Pradhan \bgroup et al.\egroup }2005]{pradhan05}
Sameer Pradhan, Kadri Hacioglu, Wayne Ward, James~H. Martin, and Daniel
  Jurafsky.
\newblock 2005.
\newblock Semantic role chunking combining complementary syntactic views.
\newblock In {\em Proceedings of the Ninth Conference on Computational Natural
  Language Learning}, pages 217--220, Ann Arbor, Michigan.

\bibitem[\protect\citename{Punyakanok \bgroup et al.\egroup
  }2008]{punyakanok08}
Vasin Punyakanok, Dan Roth, and Wen-tau Yih.
\newblock 2008.
\newblock The importance of syntactic parsing and inference in semantic role
  labeling.
\newblock {\em Computational Linguistics}, 34(2):257--287.

\bibitem[\protect\citename{Roth and Lapata}2015]{rothlapata15}
Michael Roth and Mirella Lapata.
\newblock 2015.
\newblock Context-aware frame-semantic role labeling.
\newblock {\em Transactions of the Association for Computational Linguistics},
  3:449--460.

\bibitem[\protect\citename{Roth and Woodsend}2014]{rothwoodsend14}
Michael Roth and Kristian Woodsend.
\newblock 2014.
\newblock Composition of word representations improves semantic role labelling.
\newblock In {\em Proceedings of the 2014 Conference on Empirical Methods in
  Natural Language Processing}, pages 407--413, Doha, Qatar.

\bibitem[\protect\citename{Snoek \bgroup et al.\egroup }2012]{snoek12}
Jasper Snoek, Hugo Larochelle, and Ryan~P. Adams.
\newblock 2012.
\newblock Practical bayesian optimization of machine learning algorithms.
\newblock In {\em Advances in Neural Information Processing Systems}, pages
  2951--2959, Lake Tahoe, Nevada.

\bibitem[\protect\citename{Srikumar and Roth}2013]{srikumar13}
Vivek Srikumar and Dan Roth.
\newblock 2013.
\newblock Modeling semantic relations expressed by prepositions.
\newblock {\em Transactions of the Association for Computational Linguistics},
  1:231--242.

\bibitem[\protect\citename{Tai \bgroup et al.\egroup }2015]{tai15}
Kai~Sheng Tai, Richard Socher, and Christopher~D. Manning.
\newblock 2015.
\newblock Improved semantic representations from tree-structured long
  short-term memory networks.
\newblock In {\em Proceedings of the 53rd Annual Meeting of the Association for
  Computational Linguistics and the 7th International Joint Conference on
  Natural Language Processing}, pages 1556--1566, Beijing, China.

\bibitem[\protect\citename{Toutanova \bgroup et al.\egroup }2008]{toutanova08}
Kristina Toutanova, Aria Haghighi, and Christopher Manning.
\newblock 2008.
\newblock A global joint model for semantic role labeling.
\newblock {\em Computational Linguistics}, 34(2):161--191.

\bibitem[\protect\citename{Van~der Maaten and Hinton}2008]{vandermaaten08}
Laurens Van~der Maaten and Geoffrey Hinton.
\newblock 2008.
\newblock Visualizing data using t-{SNE}.
\newblock {\em Journal of Machine Learning Research}, 9:2579--2605.

\bibitem[\protect\citename{Xiong \bgroup et al.\egroup }2012]{xiong12}
Deyi Xiong, Min Zhang, and Haizhou Li.
\newblock 2012.
\newblock Modeling the translation of predicate-argument structure for smt.
\newblock In {\em Proceedings of the 50th Annual Meeting of the Association for
  Computational Linguistics}, pages 902--911, Jeju Island, Korea.

\bibitem[\protect\citename{Xu \bgroup et al.\egroup }2015]{xu15}
Yan Xu, Lili Mou, Ge~Li, Yunchuan Chen, Hao Peng, and Zhi Jin.
\newblock 2015.
\newblock Classifying relations via long short term memory networks along
  shortest dependency paths.
\newblock In {\em Proceedings of the 2015 Conference on Empirical Methods in
  Natural Language Processing}, pages 1785--1794, Lisbon, Portugal.

\bibitem[\protect\citename{Xue and Palmer}2004]{xue04}
Nianwen Xue and Martha Palmer.
\newblock 2004.
\newblock Calibrating features for semantic role labeling.
\newblock In {\em Proceedings of the 2004 Conference on Empirical Methods in
  Natural Language Processing}, pages 88--94, Barcelona, Spain.

\bibitem[\protect\citename{Zapirain \bgroup et al.\egroup }2013]{zapirain13}
Be{\~n}at Zapirain, Eneko Agirre, Llu{\'\i}s M{\`a}rquez, and Mihai Surdeanu.
\newblock 2013.
\newblock Selectional preferences for semantic role classification.
\newblock {\em Computational Linguistics}, 39(3):631--663.

\bibitem[\protect\citename{Zhao \bgroup et al.\egroup }2009]{zhao09}
Hai Zhao, Wenliang Chen, Jun'ichi Kazama, Kiyotaka Uchimoto, and Kentaro
  Torisawa.
\newblock 2009.
\newblock Multilingual dependency learning: Exploiting rich features for
  tagging syntactic and semantic dependencies.
\newblock In {\em {Proceedings of the Thirteenth Conference on Computational
  Natural Language Learning (CoNLL 2009): Shared Task}}, pages 61--66, Boulder,
  Colorado.

\bibitem[\protect\citename{Zhou and Xu}2015]{zhou15}
Jie Zhou and Wei Xu.
\newblock 2015.
\newblock End-to-end learning of semantic role labeling using recurrent neural
  networks.
\newblock In {\em Proceedings of the 53rd Annual Meeting of the Association for
  Computational Linguistics and the 7th International Joint Conference on
  Natural Language Processing}, pages 1127--1137, Beijing, China.

\end{thebibliography}
\bibliographystyle{acl2016}

\end{document}